\documentclass[runningheads]{llncs}

\usepackage{eccv}

\usepackage{eccvabbrv}

\usepackage{graphicx}
\usepackage{booktabs}
\usepackage{tcolorbox}
\usepackage{xcolor}
\usepackage{multirow}
\usepackage{placeins}

\usepackage[accsupp]{axessibility}  

\usepackage{hyperref}

\usepackage{orcidlink}

\begin{document}

\title{Balancing Saliency and Coverage:\\ Semantic Prominence-Aware Budgeting \\for Visual Token Compression in VLMs} 
\titlerunning{PromPrune}

\author{Jaehoon Lee\inst{1} \and
Mingi Jung\inst{2} \and
Soohyuk Jang\inst{2} \and
Seungryong Yoo \inst{2} \and
Dahuin Jung \inst{3}\textsuperscript{\dag} \and
Sungroh Yoon \inst{1,2}\textsuperscript{\dag}
}

\authorrunning{J.~Lee et al.}

\institute{Interdisciplinary Program in AI, Seoul National University \and
Department of Electrical and Computer Engineering, Seoul National University \and
Department of Artificial Intelligence, Chung-Ang University \\ 
\email{jhcaptain7@gmail.com}}

\begingroup
\renewcommand\thefootnote{\dag}
\footnotetext{Corresponding authors.}
\endgroup

\maketitle

\begin{abstract}
Large Vision-Language Models (VLMs) achieve strong multimodal understanding capabilities by leveraging high-resolution visual inputs, but the resulting large number of visual tokens creates a major computational bottleneck. 
Recent work mitigates this issue through visual token compression, typically compressing tokens based on saliency, diversity, or a fixed combination of both. 
We observe that the distribution of semantic prominence varies substantially across samples, leading to different optimal trade-offs between local saliency preservation and global coverage. 
This observation suggests that applying a static compression strategy across all samples can be suboptimal. 
Motivated by this insight, we propose PromPrune, a sample-adaptive visual token selection framework composed of a semantic prominence-aware budget allocation mechanism and a two-stage token selection pipeline. Our method balances local saliency preservation and global coverage based on the semantic prominence distribution of each sample.
By allocating token budgets between locally salient regions and globally diverse regions, our method maintains strong performance even under high compression ratios. 
On LLaVA-NeXT-7B, our approach reduces FLOPs by 88\% and prefill latency by 22\% while preserving 97.5\% of the original accuracy.
Code is released at \href{https://github.com/jayaylee/PromPrune}{https://github.com/jayaylee/PromPrune}.
  
\end{abstract}

\section{Introduction}
\newcommand{\mingi}[1]{{\color{blue} #1}} 
The rapid evolution of Large Vision-Language Models (VLMs)~\cite{llava,llava_v15,llavanext,qwen2vl,qwen25vl,internvl,minicpmv} has substantially advanced multimodal understanding by embracing higher-resolution inputs and richer visual contexts, such as native-resolution processing and multi-image inputs. 
While visual tokens already significantly outnumbered text tokens in early single-image VLMs~\cite{llava,llava_v15}, recent native-resolution architectures~\cite{llavanext,qwen2vl,qwen25vl,internvl,minicpmv} further amplify this imbalance by producing substantially longer visual token sequences from high-resolution visual inputs. 
As a result, the expanded sequence length imposes considerable computational overhead, leading to higher prefill latency, increased attention computation, and larger KV-cache memory consumption.

\begin{figure}[t]
    \centering
    \includegraphics[width=\textwidth]{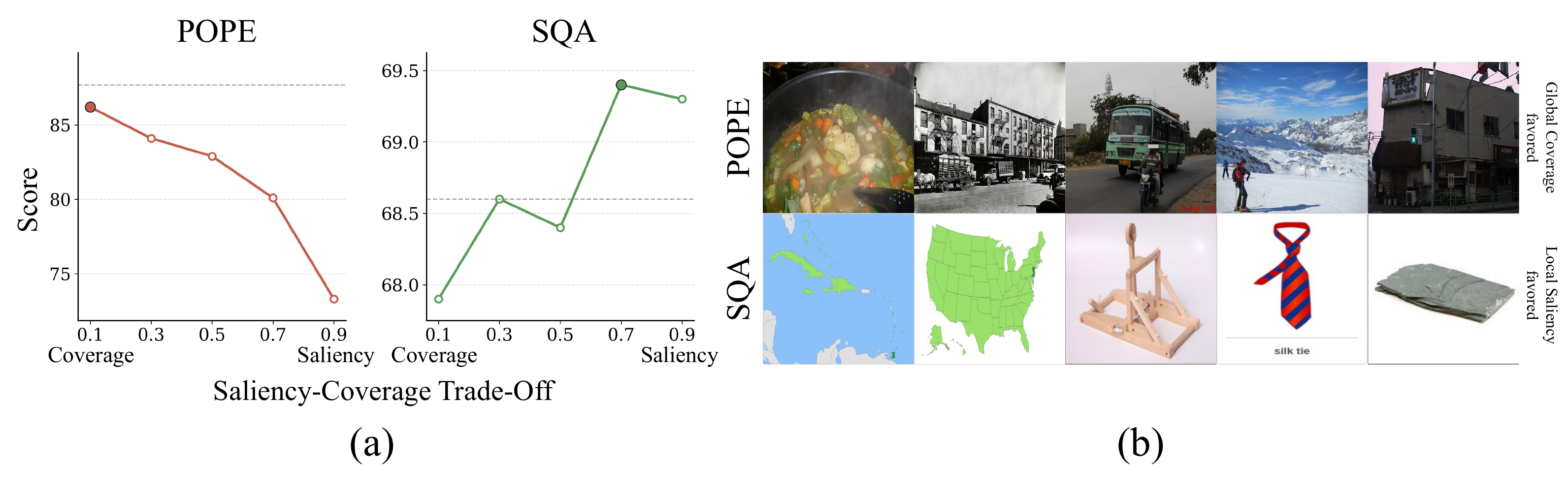}
    \vspace{-2em}
    \caption{\textbf{Saliency–Coverage Trade-Off across two distinct benchmarks.}
    (a) Performance trends of POPE and SQA under different saliency–coverage allocations with a fixed token budget. 
    (b) Example samples illustrating the contrasting semantic prominence distributions of the two benchmarks.}
    \label{fig:thumbnail}
\vspace{-2.em}
\end{figure} 
To mitigate these overheads, recent research has explored visual token compression to reduce the visual token sequence by selectively retaining the most informative visual tokens. 
Existing approaches typically prioritize a single perspective—such as \textbf{local saliency} through attention to capture prominent objects~\cite{fastv, prumerge} or \textbf{global coverage} via feature similarity to preserve broader visual context~\cite{divprune}—or enforce a static allocation between the two~\cite{vispruner}.
However, these strategies remain sample-agnostic and therefore cannot adapt to variations in how semantically informative visual cues are spatially arranged within an image.
We hypothesize that the appropriate balance between local saliency and global coverage depends on the distribution of semantically informative regions within each sample, motivating a mechanism that dynamically adjusts the balance between saliency and diversity accordingly.

To empirically validate this hypothesis, we analyze the trade-off between two complementary token selection strategies: \textbf{local saliency-driven} selection and \textbf{global coverage-driven} selection.
Specifically, under a fixed token budget, we vary the balance between local saliency-driven and coverage-driven token selection.
We conduct this study on two benchmarks that exhibit contrasting visual evidence patterns: POPE~\cite{pope}, which benefits from broad spatial context to avoid missing cues in hallucination tasks, and SQA~\cite{sqa}, which relies on localized information within a few highly salient regions for visual questioning.

As shown in Fig.~\ref{fig:thumbnail}(a), the optimal balance differs markedly across benchmarks. Performance on POPE~\cite{pope} peaks when the balance shifts toward global coverage-driven selection, whereas SQA~\cite{sqa} favors a stronger emphasis on local saliency-driven selection.
Fig.~\ref{fig:thumbnail}(b) provides qualitative examples that illustrate this difference. POPE~\cite{pope} samples often require evidence distributed across the scene to verify the presence or absence of entities, making broader coverage beneficial. In contrast, SQA~\cite{sqa} samples frequently contain concentrated informative regions where prioritizing local saliency effectively captures the most relevant visual cues without requiring extensive redundant coverage.
Together, these observations suggest that the effectiveness of visual token compression can vary depending on how informative regions are distributed within a scene. 
Consequently, fixed, sample-agnostic policies may struggle to generalize across diverse visual structures. 
This motivates a compression framework that adaptively modulates the balance between local saliency and global coverage based on the prominence distribution of each sample.

Motivated by this insight, we propose PromPrune, a sample-adaptive visual token compression framework with two key components: \textbf{Semantic Prominence-Aware Budget Allocation} and a \textbf{Two-Stage Selection Pipeline}.
Our approach estimates the distribution of semantic prominence within each sample and determines sample-adaptive saliency and coverage budgets. 
The allocated budgets are then used in a two-stage process: 
(1) selecting tokens with the highest attention scores to preserve core semantics under the saliency budget, followed by 
(2) extracting a diverse subset from the remaining candidates to maximize global coverage under the coverage budget. 
Ultimately, this framework dynamically balances salient details and diverse contextual information based on the distribution of semantic prominence, producing a highly representative fixed-length token sequence without requiring any modifications to the underlying LLM.

We extensively evaluate our approach on multiple state-of-the-art VLMs across a diverse set of vision-language benchmarks.
By dynamically tailoring the compression strategy to each sample, our method significantly reduces the visual token sequence length—thereby improving prefill latency and lowering KV-cache memory consumption—while surpassing the performance of existing static visual token compression methods.

The main contributions of our work are summarized as follows:
\begin{enumerate}
\item We identify the inherent local saliency–global coverage trade-off in visual token compression and empirically demonstrate the limitations of static token selection policies under varying semantic prominence distributions.
\item We introduce a sample-adaptive token budget allocation framework that dynamically allocates tokens between attention-based saliency and similarity-based diversity according to the semantic prominence distribution of each sample.
\item Extensive experiments demonstrate that our method reduces computational overhead and inference latency while preserving or improving vision-language understanding performance, without requiring model retraining or architectural modifications.
\end{enumerate}
\section{Related Work}
\subsubsection{Large Vision-Language Models.}
Recent Large Vision-Language Models (VLMs) typically combine a vision encoder, a large language model (LLM), and a lightweight projector that bridges the two modalities. 
Early systems demonstrated that connecting pretrained vision and language backbones enables strong multimodal understanding and instruction-following capabilities~\cite{flamingo, blip}. 
Subsequent instruction-tuned models such as LLaVA~\cite{llava} further improved generalization by training on diverse multimodal instruction datasets.
More recently, architectures have expanded to support richer visual inputs. 
Extensions such as Video-LLaVA~\cite{videollava} and LLaVA-NeXT~\cite{llavanext} introduce multi-image and high-resolution processing, while newer unified models including LLaVA-OneVision~\cite{llavaonevision} and Qwen2.5-VL~\cite{qwen25vl} aim to handle single-image, multi-image, and long-video inputs within a single framework. 
While these advances improve visual fidelity and task coverage, they also substantially increase the number and variability of visual tokens, creating significant inference-time computational and memory overhead.

\subsubsection{Visual Token Compression for VLMs.}


To alleviate the inference bottleneck caused by long visual token sequences in VLMs, recent work focuses on identifying informative tokens for visual token compression. 
Early approaches predominantly rely on saliency-driven compression, where tokens are ranked and selected according to attention-based importance scores. 
For instance, FastV~\cite{fastv} selects important tokens based on attention patterns in early LLM layers, while related methods leverage language-to-vision attention signals to identify locally salient tokens~\cite{sparsevlm}.
Another line of work focuses on diversity-based token selection. 
PruMerge~\cite{prumerge} selects and merges important tokens based on token similarity, while DivPrune~\cite{divprune} formulates token selection as a diversity-maximization problem using the Max-Min Distance Problem (MMDP)~\cite{mmdp}.
More recent methods combine local saliency preservation and global coverage promotion in hybrid strategies. 
For example, VisPruner~\cite{vispruner} retains salient tokens while supplementing them with diverse representatives.
Despite these advances, most existing approaches rely on a single selection perspective or adopt largely static policies when combining multiple criteria. 
Such sample-agnostic strategies may struggle under sample heterogeneity, where the distribution of semantic prominence varies substantially across images, motivating adaptive per-sample token allocation.
\section{Methodology}
\subsection{Preliminaries: VLM and Visual Token Compression}
Standard Vision-Language Models (VLMs)~\cite{llava, qwen25vl} encode an input image $X_v$ using a vision encoder $f_v$ and a projector $g$ to obtain a sequence of visual token embeddings
$E_v = g(f_v(X_v)) \in \mathbb{R}^{N \times d}$,
where $N$ denotes the number of visual tokens and $d$ is the LLM token embedding dimension.
These visual tokens are concatenated with textual embeddings $E_t$ and provided to a Large Language Model (LLM) $f_\phi$ for autoregressive generation.

Higher image resolution increases the number of visual tokens $N$, which directly lengthens the LLM input sequence and raises inference cost. Visual token compression mitigates this overhead by reducing the number of visual tokens to a subset of $T$ tokens $(T \ll N)$ while preserving output quality.

In this work, we compress visual tokens prior to LLM inference. This directly shortens the LLM input sequence—where most inference-time computation occurs—while remaining compatible with optimized attention kernels such as FlashAttention~\cite{flashattention,flashattention2,flashattention3}.

Formally, visual token compression can be formulated as selecting a subset $\tilde{E}_v \subseteq E_v$ that minimizes the divergence between the original and compressed output distributions of the LLM:
\begin{equation}
\tilde{E}_v^{*} = \arg\min_{\substack{\tilde{E}_v \subseteq E_v \\ |\tilde{E}_v|=T}} 
\mathcal{D} \left(
f_{\phi} \left( [\tilde{E}_v; E_t] \right),
f_{\phi} \left( [E_v; E_t] \right)
\right),
\label{eq:obj}
\end{equation}
where $\mathcal{D}$ denotes a divergence between the two output distributions, such as the KL divergence.

Directly solving Eq.~\ref{eq:obj} is combinatorial and thus intractable, so practical methods rely on surrogate signals to score or select significant tokens.
In our setting, these surrogates correspond to two complementary objectives: local saliency and global coverage.
However, existing approaches typically apply a static, sample-agnostic selection strategy, which ignores that the distribution of semantic prominence can vary substantially across samples. 

\subsection{Measuring Semantic Prominence via Spectral Entropy}\label{main:entropy}
\begin{figure}[t]
    \centering
    \includegraphics[width=0.7\textwidth]{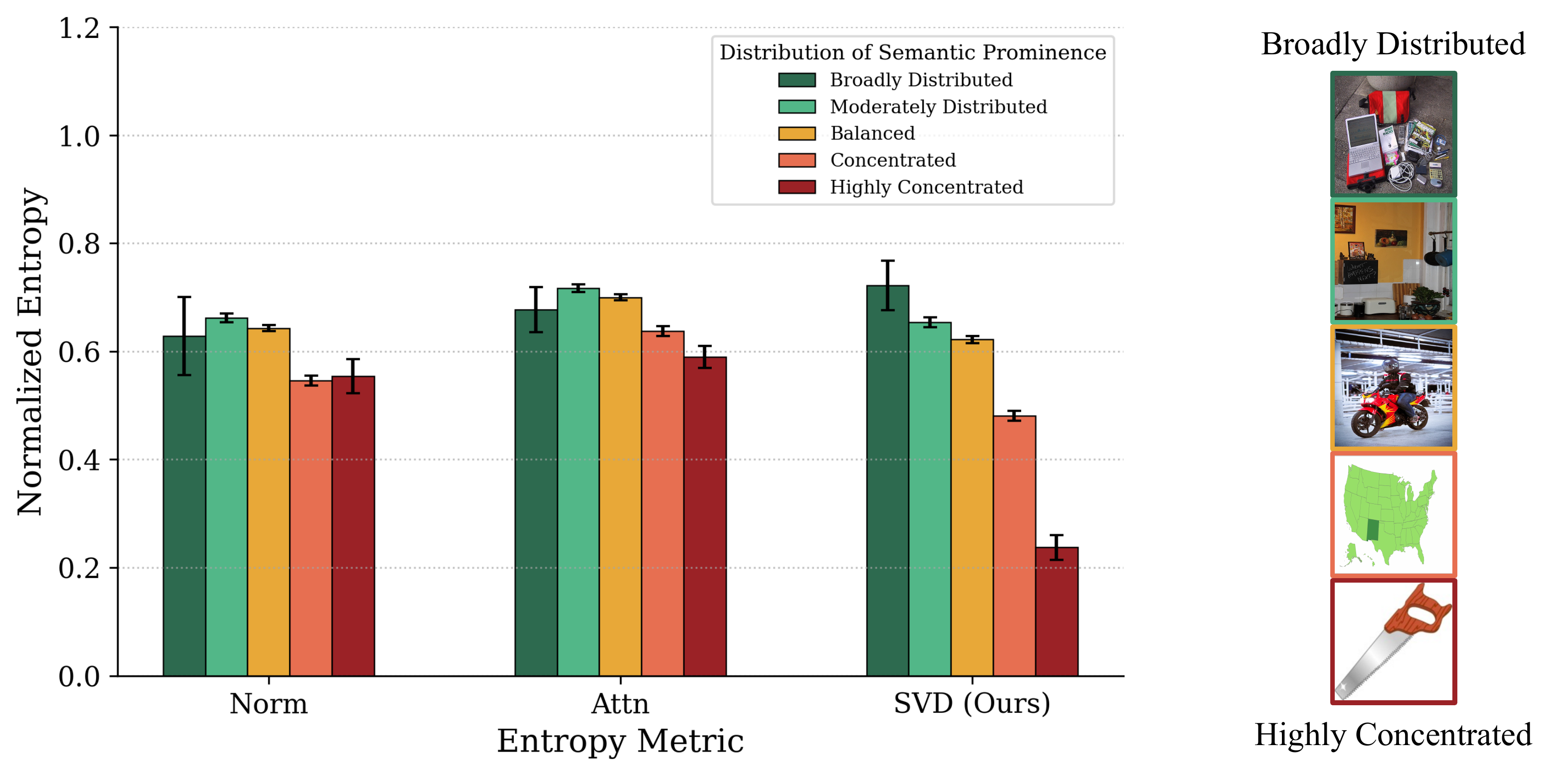}
    \caption{
    \textbf{Sensitivity of entropy metrics to semantic prominence distribution.}
Mean normalized entropy values across semantic prominence distribution levels for feature norm entropy (Norm), attention entropy (Attn), and singular value spectral entropy (Spectral).}
    \label{fig:metric}
\vspace{-2.em}
\end{figure}
\subsubsection{Limitations of Alternative Metrics.} 
To enable a sample-adaptive compression strategy, we require a robust metric that captures the distribution of semantic prominence within visual tokens. 
While standard measures such as attention entropy, computed from the entropy of attention weights across tokens~\cite{attentropy}, or feature norm entropy, derived from the entropy of token feature magnitudes~\cite{featentropy}, are intuitive candidates, whether these metrics reliably reflect the spatial distribution of semantic prominence remains an open question.

To investigate this, we conduct an empirical evaluation using samples randomly sampled from multiple image-text benchmarks, with a balanced number drawn from each benchmark.
Each sample is annotated using GPT-4o~\cite{gpt4o} with a score reflecting the spatial distribution of semantic prominence on a 1--5 scale, where lower scores indicate broadly distributed prominence and higher scores indicate more concentrated prominence.
For each prominence level, we compute the mean value of each candidate entropy-based metric across samples in that group and normalize these scores for comparison.
Additional details on the annotation process and prompting are provided in the supplementary material.

As illustrated in Fig.~\ref{fig:metric}, both feature norm entropy and attention entropy fail to exhibit consistent monotonic trends across different semantic prominence levels.
Their values overlap across scoring levels, indicating limited sensitivity to how semantic prominence is spatially distributed within the image.
This result highlights a fundamental limitation of these surrogate metrics. 
Attention entropy primarily reflects how the model distributes its internal attention across tokens, which does not necessarily correspond to the spatial distribution of semantic prominence in the scene. 
Similarly, feature norm entropy is largely driven by activation magnitudes, which can be influenced by repetitive textures or background patterns that do not necessarily reflect meaningful semantic variation. 
These observations suggest that dispersion and magnitude variation provide only partial signals of the semantic prominence distribution, motivating the need for a more principled measure of semantic prominence across visual tokens.

\subsubsection{Singular Value Spectral Entropy.}\label{sec:svd}
To quantify the distribution of semantic prominence across visual tokens, we investigate spectral entropy of the projected token feature matrix as a candidate metric. 
Inspired by prior work on effective rank analysis~\cite{rankme,trace}, we use spectral entropy as a per-sample measure of how semantic information is distributed across token representations during  inference.
Given the visual token matrix $E_v$, we compute its singular value decomposition (SVD):
\begin{equation}
E_v = U \Sigma V^\top,
\label{eq:svd}
\end{equation}
where $\Sigma = \mathrm{diag}(\sigma_1, \dots, \sigma_r)$ with $r = \min(N,d)$ and 
$\sigma_1 \ge \sigma_2 \ge \dots \ge \sigma_r \ge 0$.

Since the squared singular values correspond to the eigenvalues of $E_v^\top E_v$ and capture the variance along orthogonal principal directions,
the normalized squared singular values can be interpreted as a probability distribution over principal components. We then define the singular value spectral entropy as 
\begin{equation}
\mathcal{H}(E_v) = - \sum_{i=1}^{r} p_i \log p_i ,
\quad
p_i = \frac{\sigma_i^2}{\sum_{j=1}^{r} \sigma_j^2},    
\label{eq:spectral_entropy}
\end{equation}
where $p_i$ denotes the normalized squared $i$-th singular value. 
In practice, the singular values are efficiently computed via the eigenvalues of the Gram matrix $E_v^\top E_v$, avoiding explicit full SVD computation. 

Empirical analysis on the subset annotated with GPT-4o semantic prominence scores in Fig.~\ref{fig:metric} reveals a clear monotonic relationship: samples with broadly distributed semantic prominence exhibit higher spectral entropy, whereas images with highly concentrated prominence yield lower entropy values, indicating that spectral entropy serves as a proxy for the semantic prominence distribution across visual tokens.

Based on these observations, we employ spectral entropy as a principled measure for estimating the semantic prominence distribution of visual tokens. 
This estimate enables sample-adaptive visual token compression by allocating the fixed token budget according to the semantic prominence distribution of each input image. 

\begin{figure}[t]
    \centering
    \includegraphics[width=\textwidth]{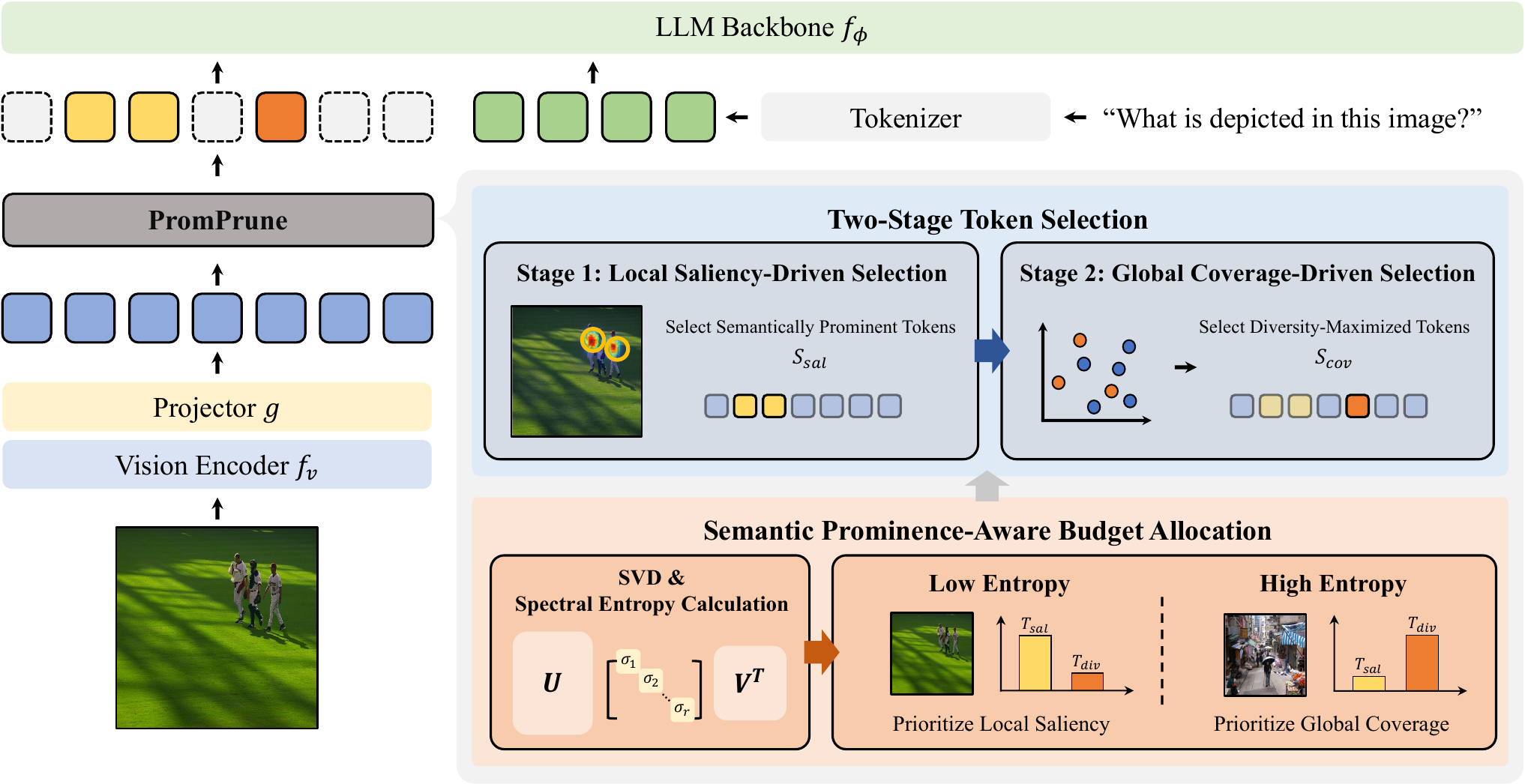}
    \caption{\textbf{Overview of the proposed PromPrune framework.}
Given an input image, the projected visual tokens are first processed by entropy-guided budget allocation to determine the saliency and coverage budgets. A two-stage token selection pipeline then retains locally salient tokens and selects diverse tokens from the unselected token pool to ensure global coverage before passing the compressed tokens to the LLM.}
    \label{fig:main_sketch}
\vspace{-1.5em}
\end{figure}

\subsection{Semantic Prominence-Aware Budget Allocation}\label{sec:budget_alloc}
While the total visual token budget $T$ is fixed, its internal allocation should adapt to the distribution of semantic prominence across visual tokens. 
Following Sec.~\ref{sec:svd}, we use spectral entropy $\mathcal{H}(E_v)$ as the signal guiding this allocation.

We decompose the total token budget into two components:
\begin{equation}
T = T_{\text{sal}} + T_{\text{cov}},
\end{equation}
where $T_{\text{sal}}$ selects locally salient tokens and $T_{\text{cov}}$ preserves global coverage.

To stabilize the allocation, we normalize spectral entropy as
\begin{equation}
\hat{\mathcal{H}}(E_v) = \frac{\mathcal{H}(E_v)}{\log r}.
\end{equation}

Low spectral entropy indicates that semantic prominence is concentrated in a small subset of tokens, where a limited number of salient tokens is sufficient to capture the dominant semantic content. 
In contrast, high spectral entropy suggests that semantic prominence is distributed across many tokens, requiring broader coverage to preserve the full visual context.

To capture this behavior, we convert the normalized spectral entropy into a coverage allocation ratio using a sigmoidal mapping:

\begin{equation}
T_{\text{cov}} =
\left\lfloor
T \cdot
\mathrm{sigmoid}\!\left(\frac{\hat{\mathcal{H}}(E_v) - \mu}{\tau}\right)
\right\rfloor,
\quad
T_{\text{sal}} = T - T_{\text{cov}},
\end{equation}
where $\mu$ and $\tau$ control the midpoint and smoothness of the transition.

This mapping produces a smooth transition between saliency-dominant and coverage-dominant regimes:

\begin{itemize}
\item \textbf{Low spectral entropy (concentrated semantic prominence):}  
$T_{\text{sal}}$ dominates while $T_{\text{cov}}$ remains small.

\item \textbf{High spectral entropy (distributed semantic prominence):}  
$T_{\text{cov}}$ increases to preserve coverage across a broader set of tokens.
\end{itemize}

By tying the allocation of the fixed budget $T$ to spectral entropy, the method adapts to the semantic prominence distribution of each input while maintaining a constant token budget across samples.

\subsection{Two-Stage Token Selection}
We adopt a two-stage pipeline that first preserves locally salient tokens and then promotes global coverage among the remaining candidates. 
The key intuition behind this design is that the most salient regions typically contain the core semantic content of the sample, regardless of how semantic prominence is distributed across tokens. 
Once these tokens are preserved, the remaining candidates mostly correspond to background or less critical regions, where the precise saliency ordering becomes less important. 

Accordingly, we first select the top $T_{\text{sal}}$ tokens according to saliency scores. 
The remaining budget $T_{\text{cov}}$ is then used to select additional tokens from the unselected token pool to improve global coverage of the final token set.

\subsubsection{Stage 1: Saliency Scoring and Budgeted Retention.} This stage preserves locally salient tokens while allowing the number of retained tokens to adapt across samples through the local saliency budget $T_{\text{sal}}$.
We compute attention-based saliency scores to rank visual tokens by their local saliency. 

Saliency scores are derived from the CLS-to-token attention in the vision encoder's self-attention map. 
Let $A^{(h)} \in \mathbb{R}^{N \times N}$ denote the attention map of head $h$ at the penultimate layer. 
The saliency score of token $i$ is defined as the CLS-to-token attention averaged across heads:
\begin{equation}
s_i =
\frac{1}{H}\sum_{h=1}^{H} A^{(h)}_{\text{CLS}, i}.
\end{equation}
Tokens are ranked according to $s_i$, and the top $T_{\text{sal}}$ tokens form the saliency token set $\mathcal{S}_{\text{sal}}$.

\subsubsection{Stage 2: Coverage Completion via Diversity Selection.}
While the saliency stage preserves locally salient tokens, it may still select multiple tokens corresponding to similar visual regions. 
The remaining token budget $T_{\text{cov}}$ is therefore used to complete the token set by selecting diverse tokens from the unselected token pool.

To encourage global coverage among the final token set, we formulate the objective as maximizing the spanned feature volume of the selected token subset in stage 2. 
This objective is naturally captured by a Determinantal Point Process (DPP)~\cite{dpp}, which provides a principled formulation for selecting mutually dissimilar tokens. 
Maximizing the determinant of the kernel matrix encourages the selected tokens to span a large feature volume.

We construct a cosine-similarity kernel over $\ell_2$-normalized token features:
\begin{equation}
L_{ij} =
\frac{e_i^\top e_j}
{\|e_i\|_2 \|e_j\|_2}.
\end{equation}
The coverage token set $\mathcal{S}_{\text{cov}}$ is obtained by selecting $T_{\text{cov}}$ tokens that maximize the determinant of the corresponding submatrix:
\begin{equation}
\mathcal{S}_{\text{cov}}
=
\operatorname*{arg\,max}_{|\mathcal{S}| = T_{\text{cov}}}
\det(L_{\mathcal{S}}).
\end{equation}
For efficiency, DPP selection is implemented using the fast greedy MAP inference algorithm~\cite{fastdpp}.
This stage complements the saliency tokens by promoting diversity among the selected token representations.

\subsubsection{Final Token Set.}
The final compressed token set is obtained by combining the outputs of both stages:
\begin{equation}
\tilde{E}_v =
\mathcal{S}_{\text{sal}} \cup \mathcal{S}_{\text{cov}},
\quad |\tilde{E}_v| = T .
\end{equation}
\section{Experimental Results}
\subsection{Experimental Setup}
\subsubsection{Models and Datasets.}
We evaluate the proposed method on multiple state-of-the-art Vision-Language Models (VLMs), including LLaVA-1.5 7B~\cite{llava_v15}, LLaVA-NeXT 7B~\cite{llavanext}, and Qwen2.5-VL 7B~\cite{qwen25vl}, to demonstrate its broad applicability across different architectures.
For LLaVA-1.5 7B, we conduct comprehensive evaluation on nine widely used image-text benchmarks, including MME~\cite{mme}, POPE~\cite{pope}, MMBench~\cite{mmbench}, TextVQA~\cite{textvqa}, SQA~\cite{sqa}, GQA~\cite{gqa}, VQAv2~\cite{vqav2}, VizWiz~\cite{vizwiz}, and MMVet~\cite{mmvet}. 
Due to the higher computational cost of evaluating newer architectures such as LLaVA-NeXT and Qwen2.5-VL, we evaluate these models on a representative subset of five benchmarks (MME, POPE, MMBench, SQA, and GQA) spanning diverse vision-language capabilities.

\subsubsection{Baselines.}
We extensively compare PromPrune against recent state-of-the-art visual token compression methods.
For LLaVA-1.5 7B, we include attention-based methods such as FastV~\cite{fastv} and SparseVLM~\cite{sparsevlm}, a diversity-based method DivPrune~\cite{divprune}, and hybrid static approaches that combine attention and diversity including VisionZip~\cite{visionzip} and VisPruner~\cite{vispruner}.
For larger models such as LLaVA-NeXT and Qwen2.5-VL, we focus on the strongest representative methods, namely the diversity-based DivPrune~\cite{divprune} and the sample-agnostic hybrid approach VisPruner~\cite{vispruner}.
\subsubsection{Implementation Details.}
All experiments are conducted using the \texttt{lmms-eval} framework~\cite{lmms_eval} under a unified evaluation setup on a single NVIDIA A40 GPU. The hyperparameters of the sigmoidal mapping ($\mu$, $\tau$) are determined from the distribution of spectral entropy values. We estimate entropy statistics from a random subset of 1,000 images sampled from the LLaVA pretraining dataset~\cite{llava} and perform a lightweight grid search around the median entropy to determine $\mu$. $\tau$ is fixed to 0.02 for all experiments. Additional implementation details are provided in the supplementary material. 
For Qwen2.5-VL~\cite{qwen25vl}, we replace the standard CLS token attention score with the global average attention score over all patch tokens, since the Qwen2.5-VL~\cite{qwen25vl} vision encoder does not include a CLS token.

\begin{table*}[t]
\centering
\caption{\textbf{Performance comparison with existing visual token compression methods on LLaVA-1.5 7B.}
Results are reported across nine image-text understanding benchmarks under different visual token budgets. \textbf{Rel.} denotes the relative performance compared to the full-token upper bound.}
\label{tab:main_llava-v15}
\small
\setlength{\tabcolsep}{4.5pt}
\renewcommand{\arraystretch}{1.12}
\resizebox{0.9\textwidth}{!}{
\begin{tabular}{lccccccccccc}
\toprule
\textbf{Method} 
& \textbf{MME} 
& \textbf{POPE} 
& \textbf{MMBench} 
& \textbf{TextVQA} 
& \textbf{SQA} 
& \textbf{GQA} 
& \textbf{VQAv2} 
& \textbf{VizWiz} 
& \textbf{MMVet} 
& \textbf{Rel.} \\
\midrule
\multicolumn{11}{c}{\textit{Upper Bound, All 576 Tokens $100.0\%$}} \\
\midrule
LLaVA-1.5 7B 
& 1507.0 & 87.7 & 63.9 & 58.1 & 68.6 & 61.9 
& 78.6 & 50.1 & 31.2 
& 100.0\% \\
\midrule
\multicolumn{11}{c}{\textit{Retain 128 Tokens \textcolor{red}{$\downarrow 77.8\%$}}} \\
\midrule
FastV~\cite{fastv} 
& 1364.2 & 67.9 & 63.1 & 55.7 & 69.4 & 54.1 
& 70.5 & 51.4 & 27.8 
& 92.0\% \\
SparseVLM~\cite{sparsevlm} 
& 1398.9 & 83.7 & 61.8 & 56.8 & 68.7 & 56.9 
& 74.9 & 49.8 & 30.0 
& 96.5\% \\
VisionZip~\cite{visionzip} 
& 1436.3 & 69.2 & 61.5 & 57.4 & 68.7 & 58.0 
& 75.5 & 51.8 & 31.5 
& 94.7\% \\
VisPruner~\cite{vispruner} 
& 1430.5 & 79.9 & 62.0 & 56.7 & 68.9 & 56.7 
& 74.9 & 52.1 & 32.9 
& 96.3\% \\
DivPrune~\cite{divprune} 
& 1359.8 & 84.4 & 59.7 & 55.3 & 68.0 & 55.6 
& 76.5 & 53.4 & 30.5 
& 96.7\% \\
\textbf{PromPrune}
& 1422.4 & 86.7 & 63.1 & 56.2 & 68.8 & 58.6 
& 76.4 & 53.5 & 33.1 
& \textbf{97.2\%} \\
\midrule
\multicolumn{11}{c}{\textit{Retain 64 Tokens \textcolor{red}{$\downarrow 88.9\%$}}} \\
\midrule
FastV~\cite{fastv} 
& 972.9 & 36.0 & 49.6 & 52.1 & 69.5 & 45.6 
& 56.0 & 48.8 & 18.9 
& 75.3\% \\
SparseVLM~\cite{sparsevlm} 
& 1189.8 & 70.3 & 57.5 & 52.7 & 68.7 & 52.3 
& 67.2 & 49.5 & 24.5 
& 88.5\% \\
VisionZip~\cite{visionzip} 
& 1365.7 & 76.5 & 59.7 & 56.0 & 68.5 & 55.4 
& 72.4 & 52.8 & 29.5 
& 94.1\% \\
VisPruner~\cite{vispruner} 
& 1359.8 & 79.9 & 59.7 & 55.3 & 68.0 & 55.6 
& 72.6 & 53.3 & 32.2 
& 94.8\% \\
DivPrune~\cite{divprune} 
& 1369.9 & 80.4 & 61.3 & 55.8 & 69.1 & 55.4 
& 74.1 & 53.5 & 28.3 
& 93.6\% \\
\textbf{PromPrune}
& 1390.9 & 85.1 & 60.7 & 55.3 & 68.7 & 57.2 
& 73.9 & 53.5 & 32.3 
& \textbf{95.3\%} \\
\midrule
\multicolumn{11}{c}{\textit{Retain 32 Tokens \textcolor{red}{$\downarrow 94.4\%$}}} \\
\midrule
FastV~\cite{fastv} 
& 885.2 & 31.7 & 38.5 & 41.9 & 43.1 & 40.8 
& 43.6 & 51.7 & 20.9 
& 62.4\% \\
SparseVLM~\cite{sparsevlm} 
& 1047.2 & 67.3 & 52.1 & 43.0 & 56.8 & 47.9 
& 58.8 & 52.0 & 18.2 
& 79.2\% \\
VisionZip~\cite{visionzip} 
& 1248.0 & 69.1 & 57.3 & 53.6 & 68.4 & 52.4 
& 67.3 & 53.1 & 25.9 
& 89.4\% \\
VisPruner~\cite{vispruner} 
& 1231.7 & 72.6 & 55.2 & 53.8 & 68.3 & 52.8 
& 67.7 & 53.1 & 31.7
& 89.8\% \\
DivPrune~\cite{divprune} 
& 1226.6 & 81.7 & 58.7 & 53.3 & 68.8 & 54.9 
& 72.2 & 53.2 & 26.3 
& 91.9\% \\
\textbf{PromPrune}
& 1325.0 & 84.4 & 59.2 & 52.6 & 68.3 & 55.2 
& 72.4 & 53.5 & 31.9 
& \textbf{93.5\%} \\
\bottomrule
\end{tabular}
}
\vspace{-2.em}
\end{table*}
\begin{table}[t]
    \centering
    \setlength{\tabcolsep}{4.5pt} 
    
    \begin{minipage}[t]{0.49\textwidth}
        \centering      
        \caption{\textbf{Performance comparison with existing visual token compression methods on LLaVA-NeXT 7B.} Results are reported across five image-text understanding benchmarks under different visual token budgets. \textbf{Rel.} denotes the relative performance compared to the full-token upper bound.}
        \label{tab:llava_next}
        \resizebox{\linewidth}{!}{
            \begin{tabular}{lcccccc}
                \toprule
                \textbf{Method} & \textbf{MME} & \textbf{POPE} & \textbf{MMB} & \textbf{SQA} & \textbf{GQA} & \textbf{Rel.} \\
                \midrule
                \multicolumn{7}{c}{\textit{Upper Bound, All 2880 Tokens $100.0\%$}} \\
                \midrule
                LLaVA-NeXT 7B & 1507.0 & 87.7 & 63.9 & 68.6 & 62.8 & 100.0 \\
                \midrule
                \multicolumn{7}{c}{\textit{Retain 640 Tokens \textcolor{red}{$\downarrow 77.8\%$}}} \\
                \midrule
                VisPruner~\cite{vispruner} & 1486.1 & 85.8 & 65.5 & 68.1 & 61.3 & 99.2\% \\
                DivPrune~\cite{divprune} & 1482.6 & 86.8 & 65.5 & 68.0 & 61.7 & \textbf{99.6\%} \\
                \textbf{PromPrune}     & 1489.8 & 87.0 & 65.0 & 68.2 & 61.8 & \textbf{99.6\%} \\
                \midrule
                \multicolumn{7}{c}{\textit{Retain 320 Tokens \textcolor{red}{$\downarrow 88.9\%$}}} \\
                \midrule
                VisPruner~\cite{vispruner} & 1435.8 & 81.6 & 63.4 & 67.7 & 58.7 & 95.9\% \\
                DivPrune~\cite{divprune} & 1438.9 & 84.6 & 63.4 & 67.4 & 61.0 & 97.7\% \\
                \textbf{PromPrune}     & 1471.0 & 85.4 & 63.1 & 67.8 & 60.7 & \textbf{97.9\%} \\
                \midrule
                \multicolumn{7}{c}{\textit{Retain 160 Tokens \textcolor{red}{$\downarrow 94.4\%$}}} \\
                \midrule
                VisPruner~\cite{vispruner} & 1315.4 & 74.0 & 58.4 & 68.3 & 55.6 & 90.6\% \\
                DivPrune~\cite{divprune} & 1349.0 & 80.4 & 61.6 & 67.3 & 59.5 & 95.0\% \\
                \textbf{PromPrune}    & 1361.3 & 80.7 & 62.3 & 67.4 & 59.1 & \textbf{95.2\%} \\
                \bottomrule
            \end{tabular}
        } 
    \end{minipage}
        \hfill 
    \begin{minipage}[t]{0.49\textwidth}
        \centering      
        \caption{\textbf{Performance comparison with existing visual token compression methods on Qwen2.5-VL 7B.} Results are reported across five image-text understanding benchmarks under different visual token budgets. \textbf{Rel.} denotes the relative performance compared to the full-token upper bound.}
        \label{tab:qwen25vl}
        \resizebox{\linewidth}{!}{
            \begin{tabular}{lcccccc}
                \toprule
                \textbf{Method} & \textbf{MME} & \textbf{POPE} & \textbf{MMB} & \textbf{SQA} & \textbf{GQA} & \textbf{Rel.} \\
                \midrule
                \multicolumn{7}{c}{\textit{Upper Bound, All 1296 Tokens $100.0\%$}} \\
                \midrule
                Qwen2.5-VL 7B & 1685.9 & 87.0 & 82.9 & 88.6 & 61.0 & 100.0 \\
                \midrule
                \multicolumn{7}{c}{\textit{Retain 512 Tokens \textcolor{red}{$\downarrow 60.5\%$}}} \\
                \midrule
                VisPruner~\cite{vispruner} & 1702.2 & 78.8 & 82.6 & 88.5 & 50.3 & 94.0\% \\
                DivPrune~\cite{divprune} & 1676.0 & 80.9 & 82.2 & 87.9 & 50.0 & 94.2\% \\
                \textbf{PromPrune}     & 1678.6 & 81.8 & 82.1 & 88.7 & 50.4 & \textbf{94.8\%} \\
                \midrule
                \multicolumn{7}{c}{\textit{Retain 256 Tokens \textcolor{red}{$\downarrow 80.2\%$}}} \\
                \midrule
                VisPruner~\cite{vispruner} & 1649.5 & 78.0 & 80.5 & 86.1 & 48.4 & 91.7\% \\
                DivPrune~\cite{divprune} & 1681.2 & 77.2 & 79.7 & 86.1 & 47.8 & 91.1\% \\
                \textbf{PromPrune}     & 1682.4 & 80.7 & 80.3 & 86.4 & 48.3 & \textbf{92.6\%} \\
                \midrule
                \multicolumn{7}{c}{\textit{Retain 128 Tokens \textcolor{red}{$\downarrow 90.1\%$}}} \\
                \midrule
                VisPruner~\cite{vispruner} & 1564.2 & 70.2 & 76.6 & 84.2 & 44.5 & 86.2\% \\
                DivPrune~\cite{divprune} & 1578.5 & 70.9 & 76.8 & 84.4 & 44.2 & 86.5\% \\
                \textbf{PromPrune}    & 1624.2 & 75.0 & 76.5 & 84.5 & 44.7 & \textbf{87.9\%} \\
                \bottomrule
            \end{tabular}
        }
    \end{minipage}  
\vspace{-1.5em}
\end{table}
\subsection{Main Results}

We evaluate PromPrune on LLaVA-1.5 7B~\cite{llava_v15} and compare it against several visual token compression baselines across nine image-text benchmarks to comprehensively assess effectiveness. Specifically, we examine how well each method preserves the model’s original capabilities when compressing the standard 576 visual tokens to strict budgets of 128, 64, and 32 tokens.

As shown in Tab.~\ref{tab:main_llava-v15}, PromPrune consistently achieves optimal results across all evaluated visual token budgets. Notably, the superiority of PromPrune becomes increasingly prominent at higher compression levels. Under the most extreme constraint of a \textbf{94.4\%} reduction in visual tokens, PromPrune successfully defends the overall model performance, maintaining \textbf{93.5\%} of the relative performance under the full-token setting. In this highly constrained setting, PromPrune outperforms VisPruner~\cite{vispruner} by a significant margin of \textbf{3.7\%p} in relative performance. These empirical results clearly demonstrate that our methodology optimally allocates the limited token budget compared to existing methods that rely on fixed or single-perspective strategies.

\subsubsection{Architectures for Higher Resolution.} 
To verify the generalization of Prom-Prune and its effectiveness in higher resolution settings where the visual token bottleneck becomes particularly severe, we conduct further experiments on advanced model architectures including LLaVA-NeXT 7B~\cite{llavanext} and Qwen2.5-VL 7B~\cite{qwen25vl}. 

LLaVA-NeXT~\cite{llavanext} generates one global image and four window crops, resulting in a massive sequence of 2880 tokens.
We evaluate PromPrune on this setting and compare it against visual token compression baselines to assess effectiveness under strict budgets of 640, 320, and 160 tokens. For Qwen2.5-VL~\cite{qwen25vl}, we resize all images to a fixed length of 1296 tokens per sample to establish an upper bound performance, and then follow the same evaluation protocol by compressing visual tokens to 512, 256, and 128 tokens.

As shown in Tabs.~\ref{tab:llava_next} and~\ref{tab:qwen25vl}, PromPrune demonstrates consistent performance improvements on both architectures. 
For LLaVA-NeXT, images are processed through a global image and multiple local window crops. 
As a result, the semantic prominence-aware allocation derived from the original image may not be fully optimal for each cropped view. 
However, to ensure fair comparison with prior work, we apply the same allocation strategy without introducing model-specific modifications. 
Adapting the allocation mechanism considering cropped views could potentially further improve the performance of our method.

In contrast, under the single-image input setting of Qwen2.5-VL, our semantic prominence-aware allocation directly reflects the visual structure of the input, leading to outperform VisPruner and DivPrune in all settings.
The performance gains become particularly pronounced under stricter token budgets, reaching up to a \textbf{1.7\%p} improvement under the 128-token budget.


\subsection{Ablation Study and Analysis}
\begin{table}[t]
\centering
\caption{\textbf{Effect of semantic prominence-aware budget allocation.} We compare the proposed semantic prominence-aware budget allocation with a fixed allocation strategy on LLaVA-1.5 7B under a 64-token budget.
The fixed allocation applies the average $T_{\text{sal}}$ of each benchmark to all samples.
Values in parentheses indicate the benchmark-wise average $T_{\text{sal}}$. \textbf{Rel.} denotes the relative performance compared to the full-token upper bound.}
\label{tab:budget}
\resizebox{0.9\linewidth}{!}{
\begin{tabular}{lcccccc}
\toprule
\textbf{Strategy} & \textbf{MMBench}(33) & \textbf{VQA$^{\text{Text}}$}(17) & \textbf{GQA}(13) & \textbf{SQA}(40) & \textbf{POPE}(12) & \textbf{Rel.} \\ \midrule
LLaVA-1.5 7B & 63.9 & 58.1 & 61.9 & 68.6 & 87.7 & 100.0\% \\ \midrule
Fixed Allocation       & 58.8               & 55.4      & 54.6               & 68.8      & 80.7            & 93.6\%          \\
\textbf{Semantic Prominence-Aware} & 60.7     & 55.3               & 57.2      & 68.7               & 85.1    & \textbf{96.1\%} \\ \bottomrule
\end{tabular}
}
\vspace{-1.5em}
\end{table}
\subsubsection{Effect of Semantic Prominence-Aware Budget Allocation.}
We evaluate the effectiveness of semantic prominence-aware budget allocation by comparing it with a fixed allocation strategy under a 64-token budget using LLaVA-1.5 7B. 
In the fixed setting, a constant $T_{\text{sal}}$ is applied to all samples, set to the average allocation produced by our method to ensure a fair comparison. 
As shown in Tab.~\ref{tab:budget}, our semantic prominence-aware allocation consistently improves performance across benchmarks, increasing the relative performance from \textbf{93.6\%} to \textbf{96.1\%}. 
These results highlight the benefit of adapting the saliency-coverage balance to the semantic prominence distribution of each input.

\begin{figure}[t]
    \centering
    \includegraphics[width=0.8\textwidth]{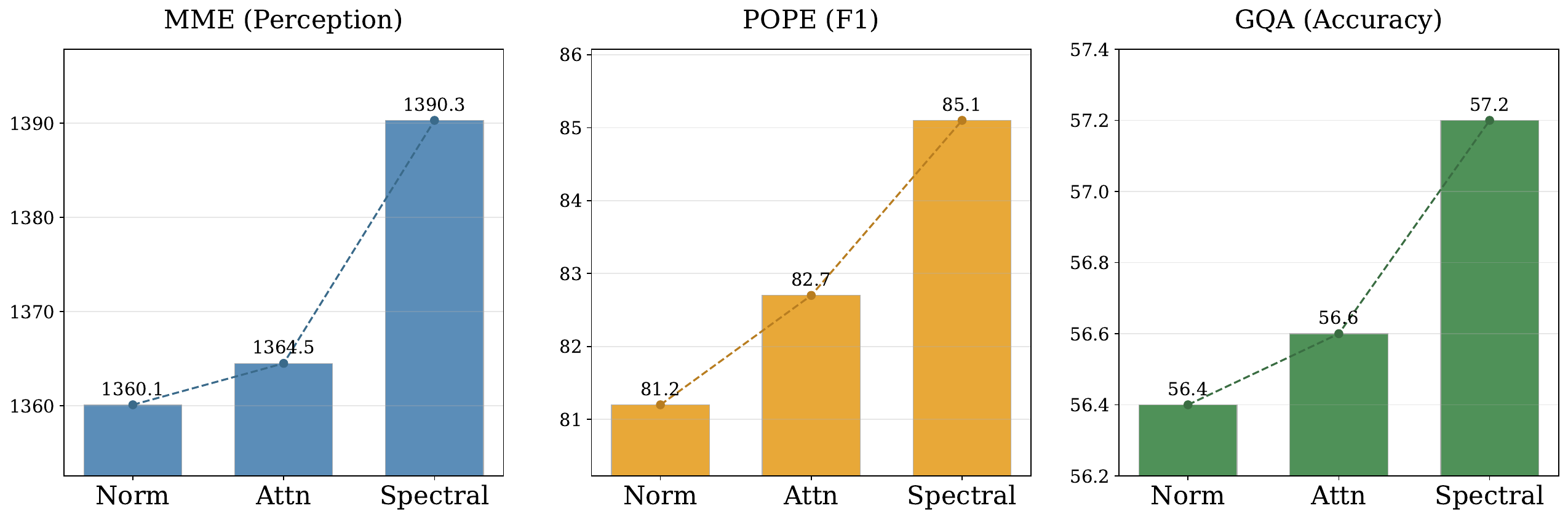}
    \caption{\textbf{Comparison of proxy metrics for estimating the distribution of semantic prominence.}
    We compare spectral entropy with two alternative proxy metrics, feature norm entropy and attention entropy on MME, POPE, GQA benchmarks.
    }
    \label{fig:ablation_metric}
\end{figure}
\subsubsection{Alternative Metrics.}
Our earlier analysis suggests that spectral entropy aligns with the distribution of semantic prominence, whereas feature norm and attention entropy exhibit notable limitations. Nevertheless, to examine the sensitivity of our approach to the choice of proxy metric, we replace spectral entropy with feature norm and attention entropy and report the resulting performance.
All methods are evaluated under the same 64-token compression setting on LLaVA-1.5 7B~\cite{llava_v15}. 
As shown in Fig~\ref{fig:ablation_metric}, spectral entropy consistently achieves the strongest performance across benchmarks. 
These results are consistent with our earlier analysis of the limitations of alternative metrics in Sec.~\ref{main:entropy}, further supporting the effectiveness of spectral entropy for estimating the distribution of semantic prominence and guiding token budget allocation.
\subsubsection{Alternative Diversity Maximization Methods.} To analyze the effect of the diversity maximization strategy, we replace the DPP-based selection with two widely used alternatives, Farthest Point Sampling (FPS)~\cite{fps} and Facility Location (FL)~\cite{fl}. FPS iteratively selects tokens that maximize the minimum distance from the selected set, promoting spatial diversity, while FL selects representative tokens by maximizing similarity-based coverage. As shown in Tab.~\ref{tab:diversity_algorithms}, while performance remains comparable across methods, DPP consistently achieves the best overall results and offers a principled formulation for globally diverse token selection.

\begin{table}[t]
\centering
\caption{\textbf{Ablation of diversity selection methods in Stage 2.} We compare different diversity maximization strategies including Farthest Point Sampling (FPS), Facility Location (FL), and Determinantal Point Process (DPP).}
\label{tab:diversity_algorithms}
\setlength{\tabcolsep}{4pt}
\resizebox{0.7\linewidth}{!}{
\begin{tabular}{lcccccc}
\toprule
\textbf{Method} & \textbf{POPE} & \textbf{MME} & \textbf{SQA} & \textbf{MMBench} & \textbf{GQA} & \textbf{Avg.} \\
\midrule
FPS & 84.4 & 1343.1 & 68.9 & 60.7 & 57.2 & 54.4 \\
FL  & 83.1 & 1377.4 & 69.1 & 60.5 & 58.0 & 54.2 \\
\textbf{DPP} & 85.1 & 1390.9 & 68.8 & 60.7 & 57.2 & \textbf{54.5} \\
\bottomrule
\end{tabular}
}
\vspace{-1.5em}
\end{table}
\begin{figure}[t]
    \centering
    \includegraphics[width=0.8\textwidth]{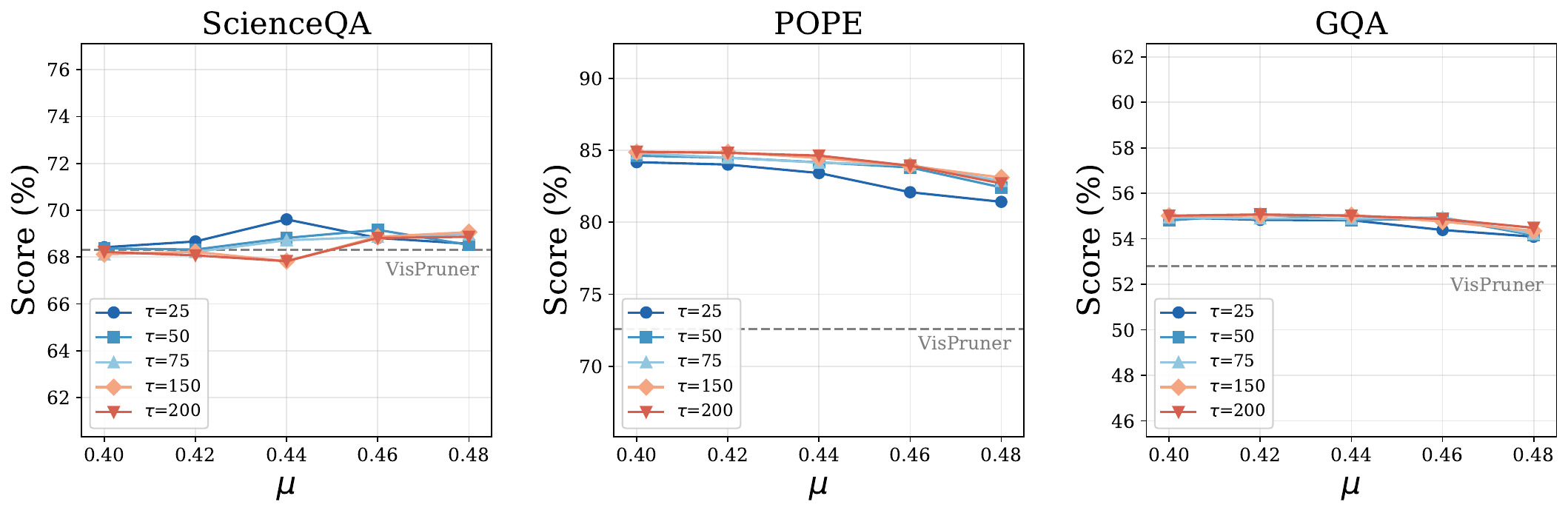}
    \caption{\textbf{Robustness of Sigmoidal Mapping Hyperparameters.} We vary $\mu$ and $\tau$ in the proposed allocation function and evaluate the resulting performance on SQA, POPE, GQA benchmarks.}
    \label{fig:robustness}
\end{figure}
\subsubsection{Hyperparameter Robustness.} To analyze the robustness of the proposed sigmoidal mapping, we vary the mapping hyperparameters across a range and evaluate the resulting performance. As shown in Fig.~\ref{fig:robustness}, performance remains stable across different values of $\mu$ and $\tau$ on all benchmarks, with only minor variations observed. Notably, the proposed method consistently outperforms VisPruner~\cite{vispruner}. These results demonstrate that the proposed sigmoidal mapping is robust to hyperparameter choices and enables more effective sample-adaptive token budget allocation than the uniform allocation strategy used in VisPruner~\cite{vispruner}.

\subsubsection{Efficiency Analysis.}
We further analyze the computational efficiency of Prom-Prune on the POPE benchmark using LLaVA-NeXT 7B~\cite{llavanext}. 
As shown in Tab.~\ref{tab:efficiency_detailed}, all pruning methods operate under the same token budget of 320 tokens and require identical FLOPs and KV cache size. 
Despite introducing additional SVD and DPP computations, our efficient implementation incurs only a modest prefill latency overhead of \textbf{14 ms} compared to DivPrune.
Nevertheless, the proposed method maintains the same computational budget while achieving the best POPE F1 score of \textbf{85.3\%} among all compared methods.
\begin{table}[t]
\centering
\caption{\textbf{Detailed efficiency analysis on LLaVA-NeXT-7B under a fixed token budget.} Efficiency metrics are measured during POPE evaluation.}
\label{tab:efficiency_detailed}
\setlength{\tabcolsep}{4pt}
\resizebox{0.8\linewidth}{!}{
\begin{tabular}{lcccccc}
\toprule
\textbf{Method} & 
\begin{tabular}[c]{@{}c@{}} \textbf{Tokens} \\ \scriptsize (Budget) \end{tabular} & 
\begin{tabular}[c]{@{}c@{}} \textbf{FLOPs} \\ \scriptsize (T) \end{tabular} & 
\begin{tabular}[c]{@{}c@{}} \textbf{Total Prefill} \\ \scriptsize (ms/sample) \end{tabular} & 
\begin{tabular}[c]{@{}c@{}} \textbf{KV Cache} \\ \scriptsize (MB) \end{tabular} & 
\begin{tabular}[c]{@{}c@{}} \textbf{Memory} \\ \scriptsize (GB) \end{tabular} & 
\begin{tabular}[c]{@{}c@{}} \textbf{POPE} \\ \scriptsize (F1) \end{tabular} \\ 
\midrule
LLaVA-NeXT-7B   & 2880 & 42.6  & 516.3  & 1440.0 & 18.62 & 87.5 \\ 
\midrule
VisPruner       & 320  & 5.02  & 321.1 & 160.0 & 16.6 & 81.6 \\
DivPrune        & 320  & 5.02  & 388.5 & 160.0 & 16.3 & 84.6 \\
\textbf{PromPrune} & 320 & 5.02 & 402.7 & 160.0 & 16.7 & 85.3 \\ 
\bottomrule
\end{tabular}
}
\vspace{-2em}
\end{table}

\section{Conclusion}
In this paper, we propose a semantic prominence-aware visual token compression framework for vision-language models. 
Our key insight is that the appropriate balance between local saliency and global coverage varies with the semantic prominence distribution of each input image.
To capture this property, we introduce spectral entropy as a proxy for semantic prominence distribution and dynamically allocate the token budget between attention-based saliency selection and diversity-based coverage preservation. 
Experiments on multiple VLMs, together with extensive ablations and analyses, demonstrate that our semantic prominence-aware compression framework consistently improves performance across diverse benchmarks by adaptively allocating the token budget based on per-sample spectral entropy.

\clearpage  


%
%
\bibliographystyle{splncs04}
\bibliography{main}

\clearpage
\appendix
\section*{Supplementary Material}
\FloatBarrier

\setcounter{figure}{0}
\renewcommand{\thefigure}{A\arabic{figure}}
\setcounter{equation}{0}
\renewcommand{\theequation}{A\arabic{equation}}

\begin{figure}[t]
    \centering
    \includegraphics[width=0.9\textwidth]{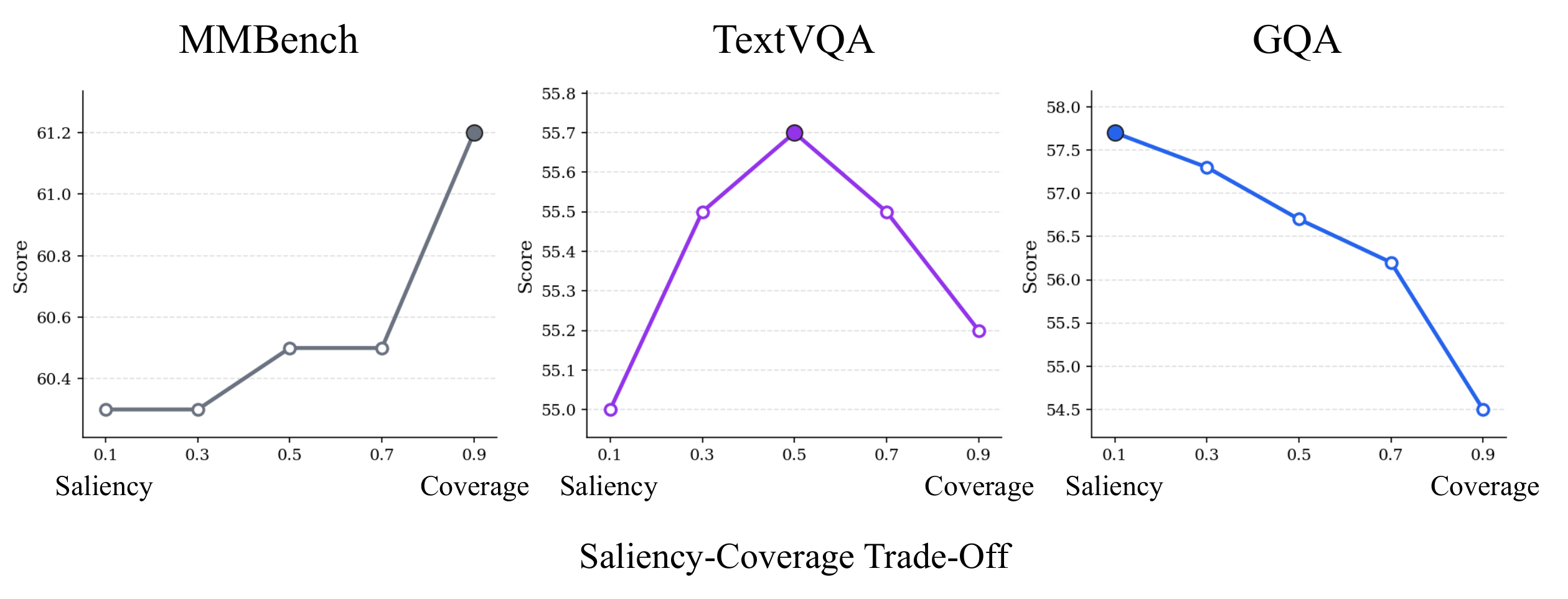}
    \caption{\textbf{Saliency--Coverage trade-off across additional benchmarks.}
    We vary the allocation ratio between saliency-driven tokens (stage~1) and coverage-driven tokens (stage~2) while keeping the total token budget fixed.}
    \label{fig:supp_tradeoff}
\end{figure}

\section{Saliency--Coverage Trade-Off Analysis}

We provide additional experimental details for the saliency--coverage trade-off analysis. 
Local salient tokens are selected using the same procedure as in the stage-1 saliency-driven selection phase of our method,
while global coverage tokens are selected using the same procedure as in the stage-2 coverage-driven selection phase.
We manually vary the salient token budget allocation ratio from 0.1 to 0.9 while keeping the total token budget fixed, and evaluate the resulting model performance across multiple benchmarks.

Fig.~\ref{fig:supp_tradeoff} reports additional results on MMBench~\cite{mmbench}, TextVQA~\cite{textvqa}, and GQA~\cite{gqa}. 
Consistent with the observations on POPE~\cite{pope} and SQA~\cite{sqa} discussed in the main paper, these benchmarks also exhibit different optimal saliency allocation ratios.

Moreover, the observed trends align with the benchmark-wise spectral entropy distributions shown in Fig.~\ref{fig:supp_benchmark_spectral} in Appendix~\ref{supp:benchmarkwise}. 
MMBench, which exhibits relatively low spectral entropy, benefits more from saliency-driven token selection. 
TextVQA, with a moderate spectral entropy distribution, favors a balanced allocation between saliency and coverage. 
In contrast, GQA shows improved performance when more tokens are allocated to global coverage, consistent with its higher entropy characteristics.

\setcounter{figure}{0}
\renewcommand{\thefigure}{B\arabic{figure}}

\begin{figure}[htbp]
\centering
\begin{tcolorbox}[
    colback=gray!5!white, 
    colframe=gray!60!black, 
    title=Prompt for Evaluating Semantic Prominence Distribution, 
    width=0.98\textwidth, 
    arc=2mm, 
    fonttitle=\bfseries\sffamily,
    boxrule=0.5pt
]
\small\sffamily

\textbf{Task Description:}\\
Rate semantic concentration in the image: how concentrated or spread the meaningful visual information is.

\vspace{1.5mm}
\textbf{Definition of a 'semantic unit' (independently meaningful entity or region):}\\
- a distinct object (person, car, animal, tool, etc.)\\
- a readable text block\\
- a meaningful object part needed for understanding an action (hands, faces, instruments)\\
- diagram elements, icons, or labeled regions

\vspace{1.5mm}
\textbf{Do NOT count:}\\
- uniform background (sky, wall, road)\\
- repeated texture without distinct objects (grass, sand, water ripples, bricks)

\vspace{1.5mm}
\textbf{Important rule:}\\
If many different objects, people, or scene elements appear across the image, the information is considered \textbf{BROADLY DISTRIBUTED} even if each object is small.

\vspace{1.5mm}
\textbf{Examples of BROAD distribution (low scores):}\\
- street scenes with many cars and people\\
- markets, crowds, sports games\\
- indoor scenes with many objects across the room\\
- natural landscapes with multiple meaningful elements (trees, buildings, animals, people)

\vspace{1.5mm}
\textbf{Examples of CONCENTRATED information (high scores):}\\
- a single object on a plain background\\
- a document or sign where the key information lies in one small area\\
- a close-up of a single person or object

\vspace{1.5mm}
\textbf{Decision procedure:}\\
1) Estimate the number of semantic units in the scene.\\
2) Estimate how widely they are spatially distributed.\\
3) If many independent objects appear across the scene, treat the image as distributed.

\vspace{1.5mm}
\textbf{Score (1--5):}\\
\textbf{5 = Highly Concentrated:} 1 main semantic unit occupying a small region.\\
\textbf{4 = Mostly Concentrated:} 1--2 units with little surrounding information.\\
\textbf{3 = Balanced:} several units but not covering the whole scene.\\
\textbf{2 = Distributed:} many objects or meaningful regions across the image.\\
\textbf{1 = Highly Distributed:} complex natural scene with many objects/interactions spread across the entire image.

\vspace{1.5mm}
\textbf{Output format:}\\
Output only one integer (1--5).
\end{tcolorbox}
\caption{\textbf{Prompt Used for GPT-4o-based Semantic Prominence Annotation.} Prompt used for GPT-4o to annotate the semantic prominence distribution of an image. The model assigns a score from 1 (highly distributed) to 5 (highly concentrated) based on the spatial concentration of semantically meaningful visual units.}
\label{fig:supp_gpt4o_prompt}
\end{figure}
\begin{figure}[t]
    \centering
    \includegraphics[width=0.9\textwidth]{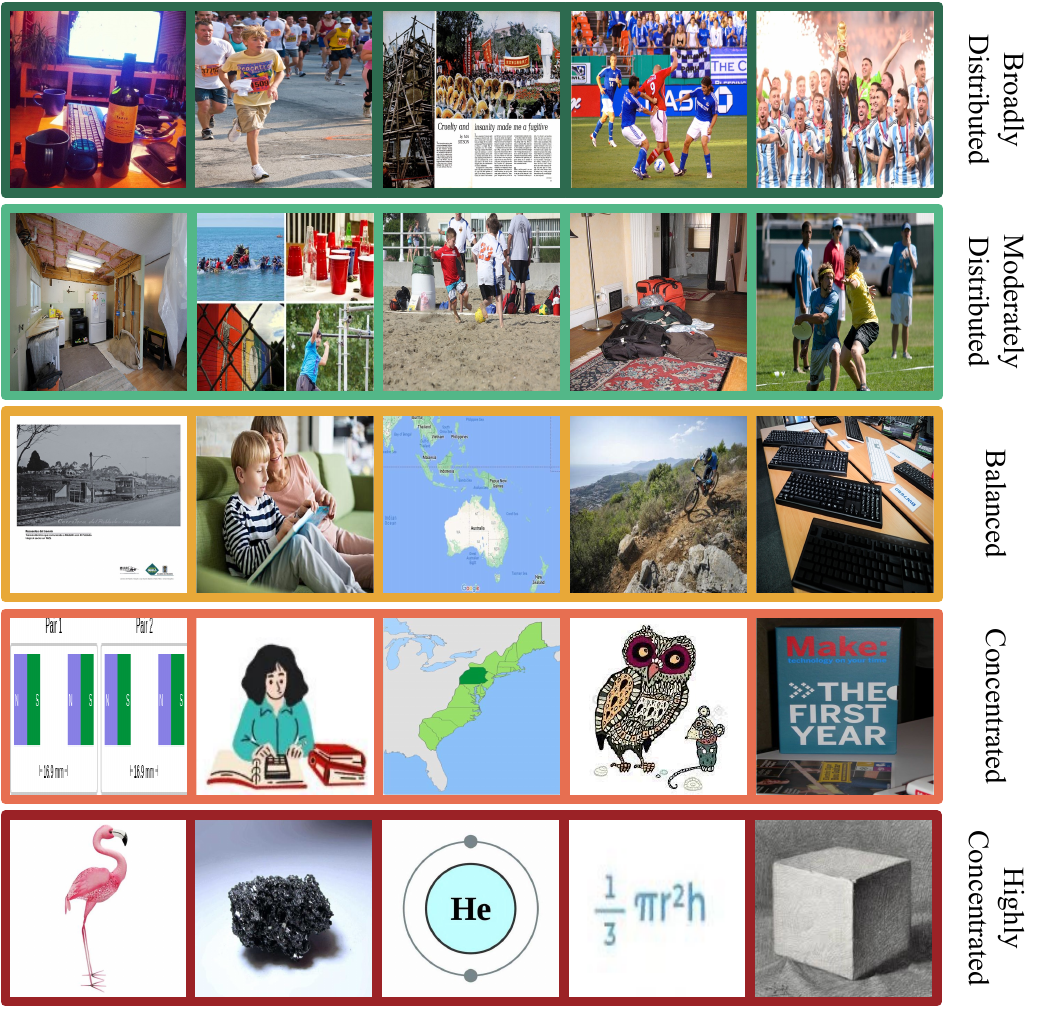}
    \caption{\textbf{Examples from different semantic prominence groups.} Representative examples from the five semantic prominence groups annotated using the protocol in Fig.~B1, ranging from broadly distributed scenes (top) to highly concentrated visual information (bottom).}
    \label{fig:supp_gpt4o_samples}
\end{figure}

\section{Empirical Validation of Proxy Metrics for Semantic Prominence Distribution}
\label{subsec:empirical_validation}

\subsection{GPT-4o-based annotation protocol}

We provide additional details for the analysis in Sec.~3.2 of the main paper.
To construct a benchmark-balanced evaluation subset, we randomly sample 100 image--text examples from each of GQA~\cite{gqa}, SQA~\cite{sqa},
POPE~\cite{pope}, MME~\cite{mme}, MM-Vet~\cite{mmvet}, MMBench~\cite{mmbench}, and TextVQA~\cite{textvqa}, resulting in a total of
700 samples.

Each sample is presented to GPT-4o~\cite{gpt4o} using the prompt shown in Fig.~\ref{fig:supp_gpt4o_prompt}.
The prompt asks GPT-4o to assign a score from 1 to 5 according to how spatially concentrated the semantically important content is in the image.

A lower score indicates that semantic prominence is broadly distributed across the scene, while a higher score indicates that the task-relevant semantic content is concentrated in a small number of localized regions.

Representative examples from the five annotation groups are shown in Fig.~\ref{fig:supp_gpt4o_samples}.
We use GPT-4o as a unified annotator across all benchmarks to ensure a consistent scoring protocol across datasets.

\subsection{Computation of candidate entropy metrics}

After obtaining the annotations, we compute three candidate entropy metrics for each sample: feature norm entropy, attention entropy, and spectral entropy.

Let $E_v = [e_1, \ldots, e_N]^\top \in \mathbb{R}^{N \times d}$ denote the projected visual token matrix, where $N$ is the number of visual tokens and $d$ is the token embedding dimension.

\subsubsection{Feature norm entropy.}
We first convert token feature magnitudes into a probability distribution by normalizing their $\ell_2$ norms:
\[
m_i = \|e_i\|_2, \qquad
q_i = \frac{m_i}{\sum_{j=1}^{N} m_j}.
\]
The feature norm entropy is then defined as
\[
\mathcal{H}_{\mathrm{norm}}(E_v)
= - \sum_{i=1}^{N} q_i \log q_i.
\]

\subsubsection{Attention entropy.}
Following~\cite{attentropy}, we compute attention entropy from the head-averaged CLS-to-token attention in the penultimate self-attention layer of the vision encoder.
\[
\alpha_i = \frac{1}{H}\sum_{h=1}^{H} A^{(h)}_{\mathrm{CLS}, i},
\qquad
p_i = \frac{\alpha_i}{\sum_{j=1}^{N} \alpha_j}.
\]
The attention entropy is defined as
\[
\mathcal{H}_{\mathrm{attn}}(E_v)
= - \sum_{i=1}^{N} p_i \log p_i.
\]
For architectures whose vision encoders lack a CLS token (e.g., Qwen2.5-VL-7B~\cite{qwen25vl}), we instead use the global average attention over patch tokens.

\subsubsection{Normalization and group-wise aggregation.}
To enable a fair comparison across metrics, we normalize each entropy by its maximum possible value.
For each GPT-4o annotation level $\ell \in \{1,2,3,4,5\}$, we compute the mean normalized entropy over all samples assigned to that level, which produces the group-wise trends shown in Fig.~2 of the main paper.

\section{Spectral Entropy-based Sigmoidal Mapping}

\subsection{Implementation Details}

We provide additional implementation details for the sigmoidal mapping used to determine the adaptive allocation between saliency and coverage tokens.
Following the implementation described in the main paper, we estimate entropy statistics from a random subset of 1,000 images sampled from publicly available datasets used in LLaVA training.
For each sample, we compute the spectral entropy of the projected visual token matrix $E_v$ as described in Sec.~3.2.

The midpoint parameter $\mu$ is initialized near the median spectral entropy of this calibration subset, and a lightweight local grid search around this initialization is performed to determine the final value.
For models using the CLIP vision encoder, e.g., LLaVA-1.5 7B~\cite{llava_v15} and LLaVA-NeXT 7B~\cite{llavanext}, the resulting value of $\mu$ is 0.42, whereas for Qwen2.5-VL-7B~\cite{qwen25vl}, whose vision encoder follows a different architecture, $\mu$ is 0.5744.
These values are fixed across all datasets and evaluation settings.
Following the main paper, the smoothness parameter $\tau$ is fixed to 0.02 in all experiments.

When PromPrune is applied to models with different visual encoders, such as Qwen2.5-VL-7B~\cite{qwen25vl}, spectral entropy is computed from the corresponding projected visual tokens using the same pipeline.
The difference in $\mu$ reflects the fact that different vision encoders produce visual token features with distinct statistical distributions.
When attention-based quantities are required, we use the global average attention over patch tokens instead of CLS-to-token attention.

\setcounter{figure}{0}
\renewcommand{\thefigure}{C\arabic{figure}}

\begin{figure}[t]
    \centering
    \includegraphics[width=0.9\textwidth]{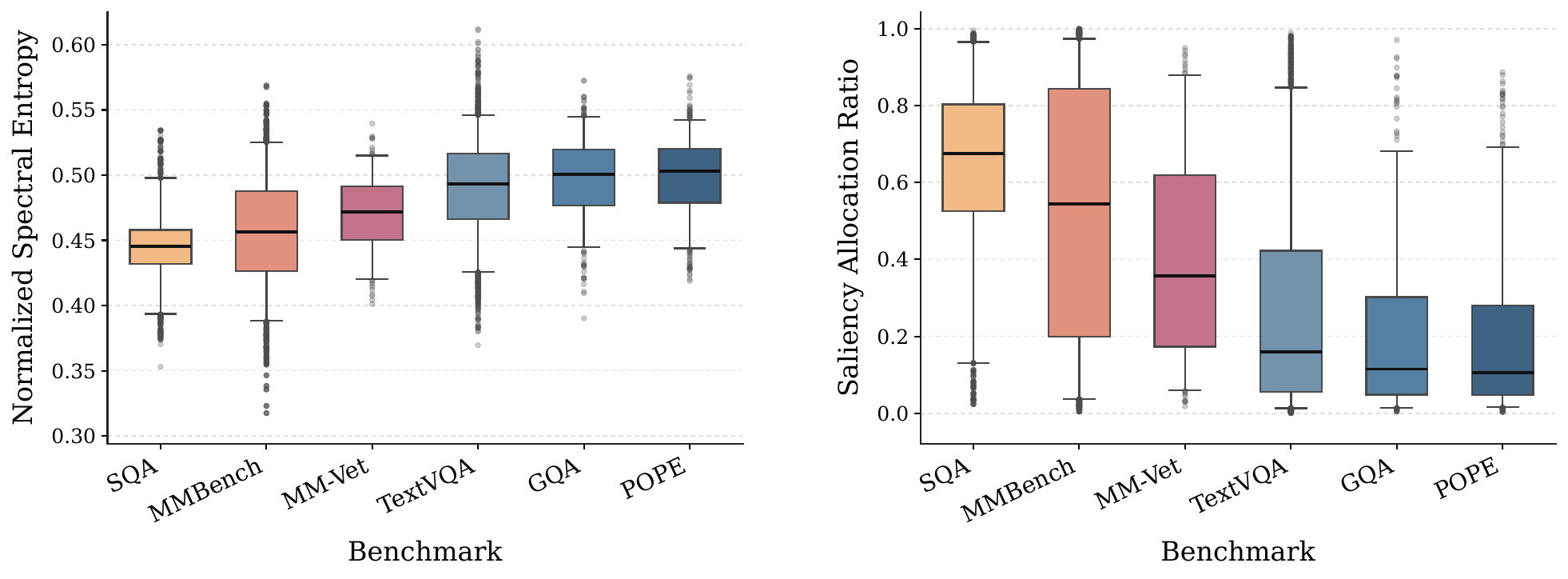}
\caption{
\textbf{The distribution of spectral entropy for different benchmarks (LLaVA-1.5).}
Benchmark-wise distributions of normalized spectral entropy (left) and the resulting saliency allocation ratio (right) computed using the CLIP vision encoder of LLaVA-1.5.
Different benchmarks exhibit different entropy distributions, which are translated into different average saliency--coverage splits by the same
sigmoidal mapping without benchmark-specific tuning.
}
    \label{fig:supp_benchmark_spectral}
\end{figure}

\subsection{Benchmark-wise Spectral Entropy and Budget Allocation}\label{supp:benchmarkwise}

Fig.~\ref{fig:supp_benchmark_spectral} illustrates the distribution of normalized spectral entropy and the resulting saliency token allocation ratio across several representative benchmarks using the CLIP vision encoder of LLaVA-1.5 7B~\cite{clip, llava_v15}.  

The left plot shows the distribution of spectral entropy values computed from visual token features, while the right plot shows the corresponding saliency allocation ratio obtained through the sigmoidal mapping described in Sec.~3.3 of the main paper.
We observe clear benchmark-level differences in spectral entropy distributions.
SQA~\cite{sqa} and MMBench~\cite{mmbench} tend to exhibit relatively lower spectral entropy, whereas GQA~\cite{gqa}, POPE~\cite{pope}, and TextVQA~\cite{textvqa} show higher entropy on average.

Through the proposed mapping, these differences are naturally translated into different average saliency allocation ratios.
Importantly, this pattern arises from the same entropy-based allocation rule without any benchmark-specific tuning.

\setcounter{figure}{0}
\renewcommand{\thefigure}{D\arabic{figure}}

\section{Ablation Study Details}

\subsection{Ablation on Semantic Prominence-Aware Budget Allocation}

For each benchmark, we hypothesize that the average number of salient tokens allocated by our sample-adaptive strategy, denoted as $T_{\text{sal}}$, approximates the optimal fixed allocation determined by the overall distribution of semantic prominence across samples.
Based on this intuition, we compare our sample-adaptive budget allocation with fixed allocation strategies under the same total token budget.

This design isolates the effect of the proposed semantic prominence-aware sample-adaptive budget allocation mechanism and allows us to evaluate whether adapting the token budget at the sample level provides advantages over globally fixed allocation strategies.

\subsection{Alternative Diversity Maximization Methods}

We describe two representative alternatives to the DPP-based diversity maximization used in PromPrune: Farthest Point Sampling (FPS)~\cite{fps} and Facility Location (FL)~\cite{fl}.  

Both methods select a subset of representative tokens from the residual visual token features, and all methods select the same number of representative tokens $M$ for a fair comparison.

Given the projected visual token feature matrix $E_v$, each token feature $e_i$ is normalized for cosine similarity:
\[
\hat e_i = \frac{e_i}{\|e_i\|_2 + \epsilon}.
\]
Given a candidate token set $R$, the goal is to select $M$ representative tokens $S \subset R$ with $|S| = M$.

\subsubsection{Farthest Point Sampling (FPS)}

FPS selects tokens that are maximally separated in the feature space.
\[
d(i,j) = 1 - \hat e_i^\top \hat e_j .
\]
Starting from an initial token, FPS iteratively selects the token whose minimum distance to the current set $S_m$ is maximized:
\[
j_{m+1}
=
\arg\max_{i \in R \setminus S_m}
\min_{j \in S_m} d(i,j).
\]
This procedure promotes a set of tokens that are widely spread in the feature space.

\subsubsection{Facility Location (FL)}

FL selects tokens that maximize coverage of the candidate set.
\[
s(i,j) = \mathrm{clip}\left(\frac{\hat e_i^\top \hat e_j + 1}{2},\, 0,\, 1\right).
\]
The facility-location objective for a subset $S$ is
\[
F(S)
=
\sum_{i \in R} w_i \max_{j \in S} s(i,j).
\]
A greedy algorithm iteratively selects the token that maximizes the marginal gain of this objective.

\setcounter{figure}{0}
\renewcommand{\thefigure}{E\arabic{figure}}

\begin{figure}[t]
    \centering
    \includegraphics[width=0.9\textwidth]{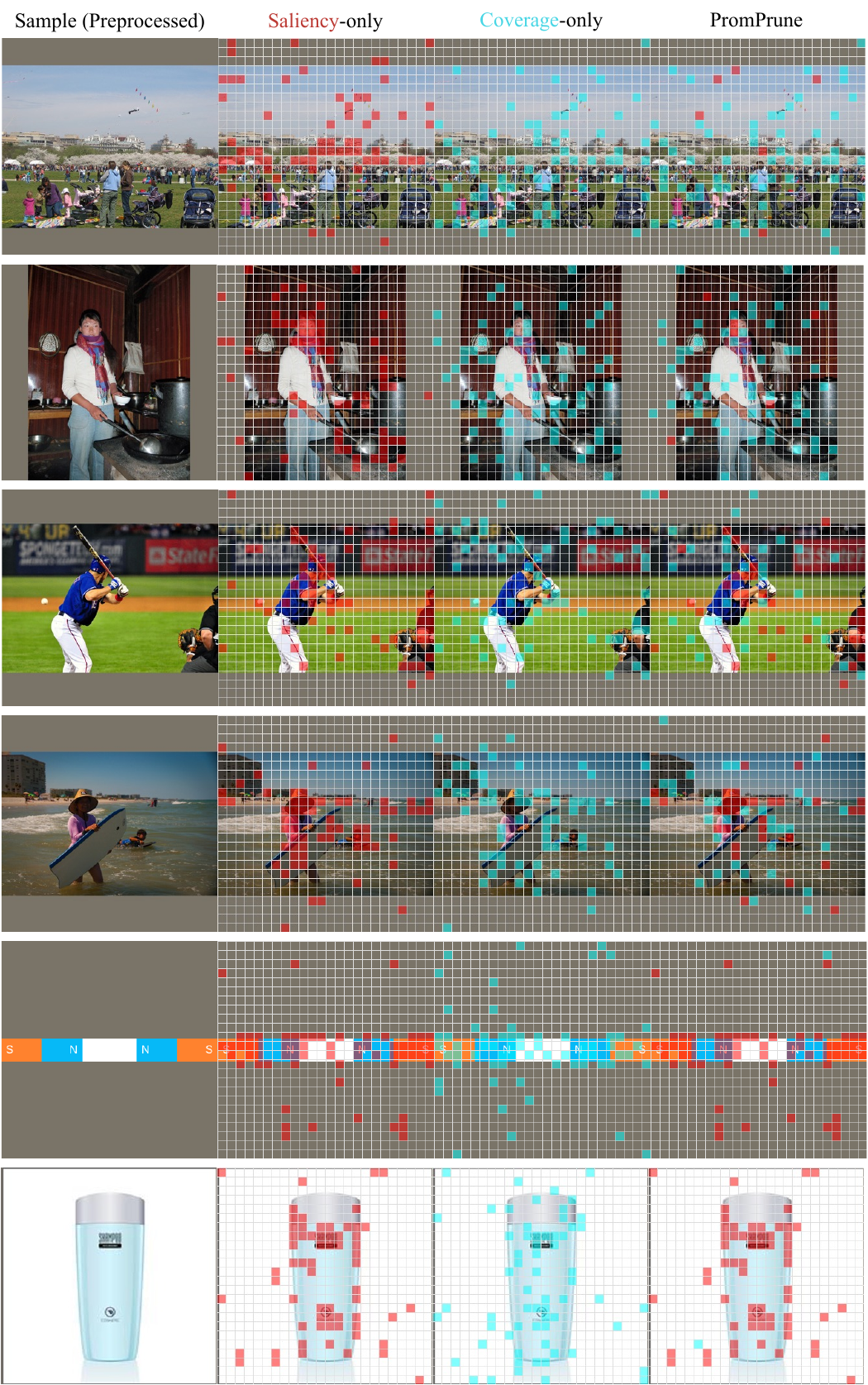}
    \caption{\textbf{Visualization of selected visual tokens ($T=64$) using LLaVA-1.5-7B.}
    Saliency-only applies stage-1 selection, Coverage-only applies stage-2 selection, and PromPrune adaptively combines both. 
    Red and cyan tokens denote saliency-driven and coverage-driven selections, respectively.}
    \label{fig:supp_selecte_tokens_viz}
\end{figure}
\section{Qualitative Comparison of Selected Tokens}

Fig.~\ref{fig:supp_selecte_tokens_viz} visualizes the visual tokens selected by different strategies when a fixed token budget of 64 is applied to LLaVA-1.5-7B. 
We compare three configurations: Saliency-only, Coverage-only, and the proposed PromPrune.

Saliency-only corresponds to applying only the stage-1 attention-based top-$K$ selection. As observed in the visualization, this strategy tends to concentrate tokens around highly salient regions, often leading to redundant selections.

Coverage-only corresponds to applying only the stage-2 diversity-driven selection, which distributes tokens more broadly across the image but may allocate tokens to regions that are less relevant.

In contrast, PromPrune adaptively allocates the token budget between the two stages based on the estimated semantic prominence distribution of each image. As a result, the selected tokens show a more balanced spatial distribution compared to the two single-stage variants.






\end{document}